\documentclass[11pt]{article}

\usepackage[final]{coling}

\usepackage{times}
\usepackage{latexsym}
\usepackage{multirow}
\usepackage{hhline}

\usepackage[T1]{fontenc}

\usepackage[utf8]{inputenc}

\usepackage{microtype}

\usepackage{inconsolata}

\usepackage{graphicx}

\usepackage{algorithm}
\usepackage{algpseudocode}
\usepackage{algorithmicx}

\usepackage{multirow}
\usepackage{xcolor}
\usepackage{color}
\usepackage{colortbl}

\usepackage{soul}
\usepackage{makecell}

\usepackage{cleveref}
\crefformat{footnote}{#2\footnotemark[#1]#3}

\title{Sinhala Transliteration: A Comparative Analysis Between Rule-based and Seq2Seq Approaches}

\author{Yomal De Mel\textsuperscript{*}, Kasun Wickramasinghe\thanks{Equal contribution.}, Nisansa de Silva \\
 Department of Computer Science \& Engineering \\ University of Moratuwa, Katubedda 10400, Sri Lanka\\
 \texttt{\{mario.23,kasunw.22,NisansaDdS\}@cse.mrt.ac.lk}\\
\textbf{Surangika Ranathunga}
\\
School of Mathematical and Computational Sciences,\\
Massey University, Auckland, New Zealand\\
\texttt{s.ranathunga@massey.ac.nz}
       }

\begin{document}
\maketitle
\begin{abstract}

Due to reasons of convenience and lack of tech literacy, transliteration (i.e., Romanizing native scripts instead of using localization tools) is eminently prevalent in the context of low-resource languages such as Sinhala, which have their own writing script.
In this study, our focus is on Romanized Sinhala transliteration. We propose two methods to address this problem:
Our baseline is a rule-based method, which is then compared against our second method where we approach the transliteration problem as a sequence-to-sequence task akin to the established Neural Machine Translation (NMT) task. For the latter, we propose a Transformer-based Encode-Decoder solution. 
We witnessed that the Transformer-based method could grab many ad-hoc patterns within the Romanized scripts compared to the rule-based method. The code base associated with this paper is available on GitHub - \url{https://github.com/kasunw22/Sinhala-Transliterator/}
\end{abstract}

\section{Introduction}

Sinhala Language, spoken by over 16 million people in Sri Lanka, presents unique challenges for computational processing due to its distinct script and structure~\cite{de2019survey}. In modern-day digital communication, it is common to use \emph{Singlish}\footnote{Not to be confused with English-based creole used in Singapore with the same name.}, where Sinhala (Sinhalese) words are written with Latin (English) script~\cite{liwera2020combination}. While the widespread use of Singlish in informal communication calls for efficient transliteration systems capable of accurately converting it into the Sinhala script, this task is made difficult by code-mixed and code-switched usage of Singlish scripts~\cite{rathnayake2022adapter, udawatta2024use}. Further, ad-hoc approximations are used by users when they approximate the \textit{Abugida} Sinhala script~\cite{liyanage2012computational} using the Latin script which is an \textit{Alphabet}~\cite{pulgram1951phoneme}. 
Yet, we do not find sufficient transliteration research done for Singlish. 

As for many NLP tasks, the early solutions for transliteration were based on rule-based techniques that relied on predefined character mappings~\cite{santaholma-2007-grammar}. However, they often struggled when confronted with the variability in the format in which  Sinhala words were written using English script~\cite{liwera2020combination}. In contrast, deep learning models, especially Transformer-based architectures~\cite{vaswani2017attention}, have proved to perform well for the transliteration task~\cite{moran-lignos-2020-effective}. However, such deep learning methods have not been used to implement Transliteration systems related to Sinhala.

This paper introduces two distinct methods, a rule-based approach and a deep learning-based approach to solve the Singlish to Sinhala transliteration problem. The deep learning based transliteration system is implemented on a pre-trained sequence-to-sequence multilingual language model, akin to a Machine Translation task. Subsequently, we evaluate their effectiveness and limitations. According to our results, we observed that the deep learning approach is more robust to language variability compared to the rule-based approach. The rest of the sections will discuss the related work, our methodology, the results we obtained, and the Conclusions.

\section{Related work}

Machine transliteration focuses on converting text from one script to another using phonetic or spelling equivalents, ideally mapping words or letters systematically between writing systems~\cite{kaur2014review}.

\subsection{Rule-based Transliteration}
Rule-based machine transliteration relies on predefined grammar rules, a lexicon, and processing software. It uses morphological, syntactic, and semantic information from source and target languages, with human experts designing rules to guide transliteration. These rules ensure the input structure and meaning are accurately mapped to the target language, preserving integrity and context in the transliterated output~\cite{kaur2014review,athukorala2024swa}.
It includes methods such as Direct Machine Translations (MT), Transfer-based MT, and Interlingual MT~\cite{sumanathilaka2023romanized}. Although effective, rule-based machine transliteration is known for being time-consuming and complex because it requires creating detailed linguistic rules to transliterate sentences from the source language to the target language~\cite{sumanathilaka2023romanized}.

\citet{tennage2018transliteration}  introduced the first transliteration system for Sinhala to English. This transliteration tool utilized character mapping tables to convert words from the native scripts of both languages into a common phonetic representation in English. The authors report that the transliteration approach allows for better preservation of word ordering and more accurate transliteration of phrases. 
Their system shows a good accuracy for handling of loanwords—where both languages share similar transliterated forms—and also enhances the overall translation quality by allowing for better mapping of linguistic structures, thus addressing the challenges posed by the morphological richness of both languages.

Hybrid transliteration systems that combine rule-based methods with a trigram model have shown to improve the accuracy of converting Singlish to Sinhala~\cite{liwera2020combination}. The rule-based component applies predefined rules for vowels and consonants, while the trigram model uses statistical patterns from social media comments to address the variability and ambiguity of Singlish input.

\subsection{Transformers for multilingual Sequence-to-Sequence Generation Tasks}
For sequence-to-sequence (Seq2Seq) generation tasks such as Machine Translation (MT), the proven architecture is the Encoder-Decoder architecture. When it comes to multilingual Transformer-based pre-trained Encoder-Decoder architectures,  mT5~\cite{xue-etal-2021-mt5} which is based on T5~\cite{raffel2020exploring}, mBART~\cite{liu2020multilingual} which is based on BART~\cite{lewis2019bart}, M2M100~\cite{fan2020beyond}, MarianNMT~\cite{tambouratzis-2021-alignment} have been popular choices. The advantage of the Transformer-based Encoder-Decoder architecture is that due to its self-attention and cross-attention mechanisms, the relationships with and among the source and the target sequence are properly captured~\cite{vaswani2017attention}. Seq2Seq, has since been utilized in domains other than MT~\cite{de2020legal}.   

\subsection{Translation Models with Sinhala Language Support}
There are several free and open-source multilingual translation models that include Sinhala. Among them mT5, mBART, M2M100, MarianNMT, and NLLB\footnote{\url{https://github.com/facebookresearch/fairseq/tree/nllb?tab=readme-ov-file}}\cite{costa2022no} are prominent. Both M2M100 and NLLB use the same model architecture but two different training datasets. M2M100 uses CCMatrix~\cite{schwenk-etal-2021-ccmatrix} and CCAlighned~\cite{el2019massive} datasets while NLLB uses the NLLB~\cite{costa2022no} dataset. On the other hand, MarianNMT model uses a different encoder-decoder architecture, and the dataset they use is OPUS-100~\cite{zhang-etal-2020-improving}. Both mBART and mT5 have been used for various Sinhala text generation tasks, including Machine Translation~\cite{niyarepola2022math, ranathunga2024sitse, thillainathan2021fine, lee2022pre}. However, according to a recent study by~\citet{ranathunga-etal-2024-quality}, NLLB has proven to be the best among them for translation tasks that involve Sinhala.

\subsection{Deep Learning based Transliteration}
\citet{deselaers-etal-2009-deep} proposed a deep belief system-based transliteration solution using Deep Belief Networks (DBN). DBN architecture is almost similar to the encoder-decoder architecture. \citet{deselaers-etal-2009-deep} mentioned that transliteration can be considered a translation task at the character level. Subsequent neural network-based (NN) solutions for the transliteration task mainly relied on recurrent models such as simple RNN, LSTM, and GRU~\cite{shao-nivre-2016-applying,7985375,rosca2016sequence,kundu-etal-2018-deep}.
\citet{zohrabi-etal-2023-borderless} have used a Transformer-based approach for the transliteration of Azerbaijani.
A comparative evaluation of LSTM, biLSTM, GRU, and Transformer architectures for named entity transliteration has been carried out by \citet{moran-lignos-2020-effective}. According to their evaluation, Transformer-based encoder-decoder architectures outperform other architectures.

\section{Methodology}

\subsection{Rule-Based Transliteration System}
Our rule-based approach uses predefined linguistic rules to map Latin script (Singlish) to Sinhala script. These rules
cover vowels, consonants, diacritics, and special characters. It extends the rule-based transliteration system of~\citet{tennage2018transliteration} with a few additions to the mapping rules when considering two and three-character mapping. Some of the rules defined are shown in Table~\ref{tab:transliteration_rules}, where newly added rules are highlighted. The process involves two primary stages: rule definition and application.

\begin{table}[!htb]
\centering

\includegraphics[width=0.5\textwidth]{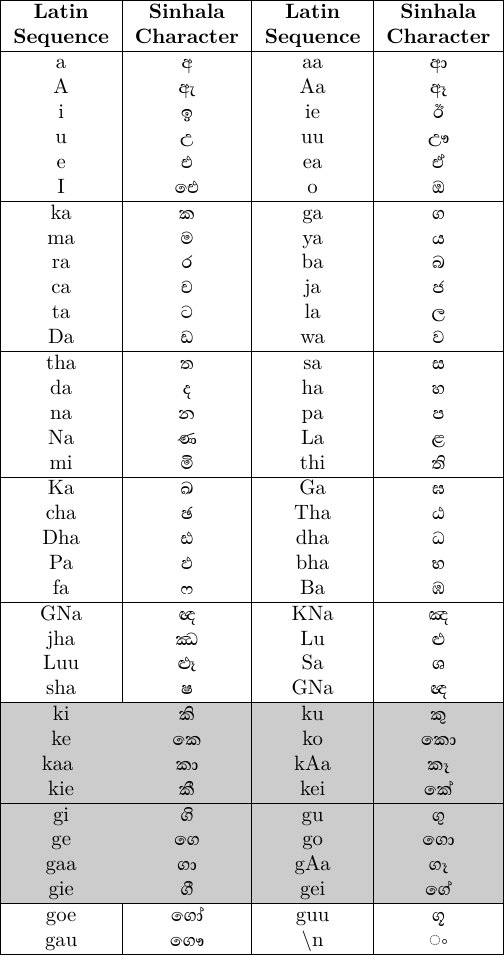}

\caption{Transliteration rules. The highlighted rules were added by us.}
\label{tab:transliteration_rules}

\end{table}

\begin{algorithm}[!htb] 
\caption{Transliteration Algorithm}
\label{alg:transliterate}
\begin{algorithmic}[1]
\Require{Latin script word \texttt{word}} 
\Ensure Sinhala script word
\State \texttt{result} $\gets$ \texttt{``''} \Comment{Initialize an empty string}
\State \texttt{i} $\gets$ 0 \Comment{Initialize index}
\While{\texttt{i} $<$ \texttt{length(word)}}
    \State \texttt{matched} $\gets$ \textbf{False}
    \For{\texttt{length} in \{3, 2, 1\}} \Comment{Check substrings of decreasing length}
        \State \texttt{substring} $\gets$ \texttt{word[i:i + length]}
        \If{\texttt{substring} \textbf{in} \texttt{transliteration\_table}}
            \State \texttt{result} $\gets$ \texttt{result + transliteration\_table[substring]}
            \State \texttt{i} $\gets$ \texttt{i + length}
            \State \texttt{matched} $\gets$ \textbf{True}
            \State \textbf{break}
        \EndIf
    \EndFor
    \If{\textbf{not} \texttt{matched}}
        \State \texttt{result} $\gets$ \texttt{result + word[i]}
        \State \texttt{i} $\gets$ \texttt{i + 1}
    \EndIf
\EndWhile
\State \Return \texttt{result}
\end{algorithmic}
\end{algorithm}

The transliteration function processes each input word and converts it to Sinhala using a character-by-character matching strategy, as detailed below.  The pseudocode is shown in Algorithm~\ref{alg:transliterate}.

\begin{itemize}
    \item \textbf{Input Processing:} The system reads the input word in Latin script and ensures it contains only Latin characters.
    \item \textbf{Longest Match Strategy:} For each character sequence, the system matches the longest possible substring (up to three characters). This ensures that multi-character sequences such as ``th'' or ``aa'' are mapped correctly before shorter, single-character matches.
    \item \textbf{Rule Application:} If a match is found in the transliteration table, the corresponding Sinhala character is appended to the result. If no match is found, the character is added as is.
    \item \textbf{Output Generation:} The transliterated word is returned and added to the output dataset.
\end{itemize}

\subsection{Deep Learning-Based Transliteration System}

 In this approach, we model transliteration as a translation task, as suggested by ~\citet{deselaers-etal-2009-deep}. Even though decoder-only Large Language Models (LLMs) are the state-of-the-art choice for most of the NLP tasks including Machine Translation nowadays, for many \emph{low-resource language} translation tasks, still sequence-to-sequence modes are commonly used~\cite{ranathunga2023neural}. Considering these factors, a Transformer-based encoder-decoder model is our second approach to solving the reverse transliteration problem.

Apart from the context-based generation, another advantage of this approach is that unlike in rule-based approaches, we do not need to manually define the rules and we only need to find or create a rich dataset that covers the possible scenarios that could occur during the inference time. Moreover, the code-mixed and code-switched cases can also be easily addressed in this approach simply by extending the training dataset accordingly.

To have better accuracy, rather than training the model from scratch, we used an existing multilingual pre-trained sequence-to-sequence model that is trained for the translation task, which has coverage for Sinhala as well. To be specific, we have selected the 418M version of the M2M100 model\footnote{\label{M2M100}\url{https://huggingface.co/facebook/m2m100_418M}} \cite{fan2020beyond} as our base model and fine-tuned it for Romanized-Sinhala and Sinhala as a translation pair. We used the existing English language code (i.e. $en$) for Romanized Sinhala and the Sinhala language code (i.e. $si$) for Sinhala. The reason for selecting M2M100 is that the MarianMT translation quality for the Sinhala-English pair is a bit worse than M2M100 and NLLB models (see Table~\ref{tab:translation_model_comparison}). Both NLLB and M2M100 use the same model architectures and the translation qualities are almost similar (Table~\ref{tab:translation_model_comparison}). We choose M2M100 over NLLB since NLLB model weights are bound with some additional restricted terms and conditions\footnote{\url{https://github.com/facebookresearch/fairseq/blob/nllb/LICENSE.model.md}} while M2M100 weights are not\footnote{\url{https://choosealicense.com/licenses/mit/}}.

We fine-tuned M2M100 model in a way that the Romanized script is considered as the English translation of the corresponding Sinhala script. We used the M2M100 model's tokenizer\cref{M2M100} for the tokenization process. Since the model already knows the basic linguistics from the translation task, it only needs to learn the relationship between the two new language pairs. Also in Romanized typing, it is more common to use code-mixed usage within the content. Furthermore, since we are using a Transformer-based model, the context is also taken into account when the transliteration is done.

\newcommand{\dgc}{0.8}
\newcommand{\lgc}{0.9}
\definecolor{dg}{rgb}{\dgc, \dgc, \dgc}
\definecolor{lg}{rgb}{\lgc, \lgc, \lgc}
\newcommand{\worse}{\cellcolor[gray]{\dgc}}
\newcommand{\bad}{\cellcolor[gray]{\lgc}}
\newcommand{\capBox}[2]{%
  \begingroup\setlength{\fboxsep}{0.8pt}%
  \colorbox{#2}{\texttt{\hspace*{0.5pt}\vphantom{Ay}\footnotesize #1\hspace*{0.5pt}}}%
  \endgroup
}

\begin{table*}[!htb]
\centering
\includegraphics[width=\textwidth]{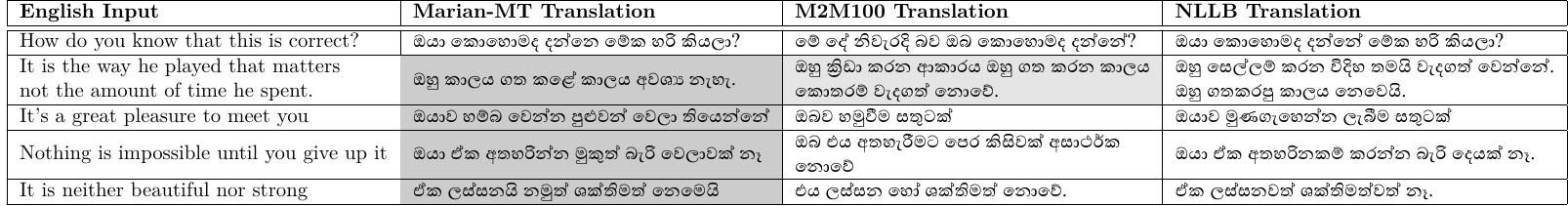}
\caption{Qualitative evaluation of translation models. Records shaded in \capBox{\strut light gray}{lg} indicate the translations are slightly incorrect and the \capBox{\strut dark gray}{dg} shaded ones are really bad translations. Non-shaded ones are correct translations.}
\label{tab:translation_model_comparison}

\end{table*}

\section{Implementation}
\subsection{Dataset Preparation} \label{sec:dl_dataset_creation}
The task is a sequence-to-sequence text generation task, specifically developing a reverse transliterator that converts Romanized Indo-Aryan languages to their native scripts. Therefore what we need is a parallel dataset that contains Romanized text and the corresponding native script.

In order to create the training dataset, we used the Dakshina~\cite{roark-etal-2020-processing} and Swa-Bhasha~\cite{sumanathilaka2023swa, sumanathilaka2024swa} datasets. We further augmented the datasets by adding some ad-hoc nature to the Romanized scripts by removing vowels and applying different common typing patterns. See Table~\ref{tab:data_augmentation} for examples. We created a dataset consisting of 10k parallel data points using these data sources. We split that into a training set of 9k data points and a validation set of 1k data points for the model training and validation. 

We have evaluated our two approaches on the test sets\footnote{\url{https://github.com/IndoNLP-Workshop/IndoNLP-2025-Shared-Task}} provided by the shared task on "Reverse Transliteration on Romanized Indo-Aryan languages using ad-hoc transliterals", organized by the IndoNLP workshop with COLING 2025. Test set 1 consists of 10,000 parallel entries containing general Romanized typing patterns and, test set 2 consists of 5000 parallel entries with ad-hoc Romanized typing patterns that come across in practical scenarios making it very challenging to solve the reverse transliteration task. The original datasets were not well structured. Therefore we converted these datasets into CSV format, containing Romanized Sinhala script (Singlish) sentences in one column and the corresponding expected Sinhala script in another.

\begin{table}[!htb]
\centering

\includegraphics[width=0.5\textwidth]{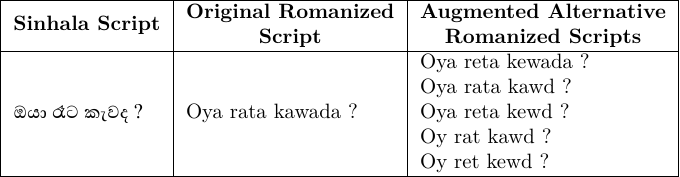}

\caption{Data augmentation example}
\label{tab:data_augmentation}

\end{table}

\subsection{Computational Resources}
We used an NVIDIA Tesla T4 16GB GPU for the training process. The important training hyper-parameters have been listed in Table~\ref{tab:training_params}.

\begin{table}[!ht]
    \centering
    \resizebox{\columnwidth}{!}{
        \begin{tabular}{p{0.75\linewidth}p{0.25\linewidth}}
        \hline
        Hyperparameter & Value\\ \hline
        learning rate & 2e-5 \\
        epochs & 3  \\
        train batch size & 8 \\
        gradient accumulation steps & 1 \\
        effective training batch size & 8 \\
        training precision & fp16 \\
        weight decay & 0.01 \\
        optimizer & Adam \\
        learning rate scheduler & linear \\
        training dataset & 9000\\
        evaluation dataset & 1000\\ \hline
        \end{tabular}
    }
\caption{Training hyper-parameters of the deep learning model}
    \label{tab:training_params}
\end{table}

\subsection{Evaluation Metrics}

To assess the accuracy of the transliteration, we use three key metrics:
\begin{itemize}
    \item \textbf{Word Error Rate (WER):} Measures the difference between the predicted and reference sentences at the word level. The lower the WER the better.
    \item \textbf{Character Error Rate (CER):} Evaluates character-level accuracy by calculating the number of edits needed to convert the predicted output to the reference. The lower the CER the better.
    \item \textbf{BLEU Score:} Assesses the overlap between predicted and reference outputs. The higher the BLEU score the better.
\end{itemize}

We used the metric implementations of Python \texttt{evaluate}\footnote{\url{https://huggingface.co/docs/evaluate/v0.4.0/en/index}} library for our evaluation.

\section{Results and Discussion} \label{sec:results_and_discussion}

\begin{table}[!ht]
    \centering
    \resizebox{\columnwidth}{!}{
        \begin{tabular}{llll}
        \hline
        Approach & \makecell{Evaluation\\Matrix} & \makecell{Average Result\\for Test Set 01} & \makecell{Average Result\\for Test Set 02} \\ \hline
        \multirow{3}{*}{Rule-based} & WER & 0.6689 & 0.6809 \\
        & CER & 0.2119 & 0.2202 \\
        & BLEU & 0.0177 & 0.0163 \\ \hline
        \multirow{3}{*}{DL-based} & WER & \textbf{0.1983} & \textbf{0.2413} \\
        & CER & \textbf{0.0579} & \textbf{0.0789} \\
        & BLEU & \textbf{0.5268} & \textbf{0.4384} \\ \hline
        \end{tabular}
    }
\caption{Results for rule-based and deep learning based techniques}
    \label{tab:both_approach_results}
\end{table}

Table~\ref{tab:both_approach_results} shows the evaluation metrics for rule-based and deep learning-based approaches evaluated on the provided two test sets. 
As can be seen in Table~\ref{tab:compare_robustness}, the deep learning approach is more robust to the ad-hoc variations of Romanized typing compared to the rule-based approach.

\begin{table}[!htb] 
    \centering
    
    \includegraphics[width=0.5\textwidth]{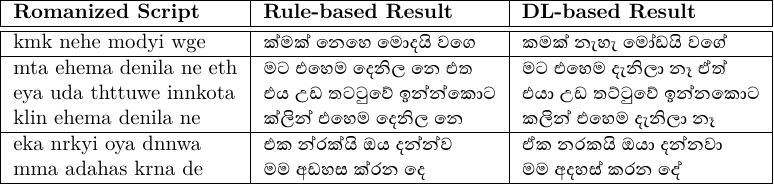} 
    \caption{Robustness comparison of two approaches}
    \label{tab:compare_robustness}
\end{table}

Nevertheless, the efficiency concerned, the rule-based approach is much faster than the deep learning approach. In the CPU, the deep learning approach becomes extremely slow making it hard to use for real-time applications. In contrast on a GPU, we can achieve real-time performance for the deep learning approach as well. Check Table~\ref{tab:compare_performance} for the  results related to computing efficiency. We used output \emph{tokens per second} (TPS) as the performance measure. According to Table~\ref{tab:compare_performance}, we can expect better performance values for the deep learning approach with lower precision setups (i.e. fp16, INT8, INT4, etc.) possibly with a slight accuracy compromisation. 

\begin{table}[!ht]
    \centering
    \resizebox{\columnwidth}{!}{
        \begin{tabular}{p{0.25\linewidth}p{0.25\linewidth}p{0.25\linewidth}p{0.25\linewidth}}
        \hline
        \multirow{2}{*}{Rule-based} & \multicolumn{3}{c}{deep learning}   \\
        \hhline{~---}
         & CPU (fp32) & GPU (fp32) & GPU (fp16) \\        
        \hline
        $>$200,000 & $\sim$3 & $\sim$35 & $\sim$65 \\ \hline
        \end{tabular}
    }
\caption{Speed (in TPS) comparison of the two approaches.}
    \label{tab:compare_performance}
\end{table}

\section{Conclusion}
We have experimented with two approaches for the transliteration task for Romanized Sinhala and English. The first approach is a rule-based statistical approach. The second approach addresses the transliteration task as a translation task using a pre-trained multilingual encoder-decoder language model. Both approaches have their own pros and cons. When it comes to accuracy, the deep learning approach outperformed the rule-based method while in terms of efficiency, it is the other way around.

\section*{Limitations}
The deep learning-based approach does come with a compromise of efficiency to the accuracy. The quality of the output of the deep learning approach heavily depends on the quality of the training data. 

The rule-based transliteration system for converting Latin script to Sinhala faces several key challenges. A primary limitation is ambiguity handling: certain Latin character sequences can map to multiple Sinhala characters depending on context. Without contextual awareness, the system processes each character sequence independently, leading to inaccuracies, especially with complex or compound words where pronunciation depends on neighbouring syllables. 

Additionally, users often spell the same word differently based on their typing preferences or ease. For instance, the Romanized term ``mama'' could correspond to different Sinhala words such as \includegraphics[height=1.5\fontcharht\font`\B,trim=0 1mm 0 -1mm]{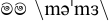} (Nominative \textit{I}), \includegraphics[height=1.5\fontcharht\font`\B,trim=0 1mm 0 -1mm]{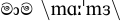} (Accusative \textit{specifically me}), or \includegraphics[height=1.5\fontcharht\font`\B,trim=0 1mm 0 -1mm]{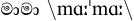}  (Nominative \textit{uncle}). 

This inconsistency introduces ambiguity, making it difficult to define rigid transliteration rules. In contrast, deep learning models can better handle such variations by learning context and patterns from large datasets, offering more flexibility and accuracy.

Additionally, the predefined rules may not cover all linguistic nuances, resulting in errors when encountering words that deviate from standard structures. Morphological complexities, such as inflections or compound words, further challenge the system, as it does not account for grammatical context.

We have used a training set of 9k parallel entries for the deep-learning model fine-tuning. Having an extended training set covering more practical cases could lead to better results.

As future work, we plan to address these limitations and also experiment with LLMs for the transliteration task.

\bibliography{custom}

\end{document}